# Flavonoid Fusion: Creating a Knowledge Graph to Unveil the Interplay Between Food and Health

Aryan Singh Dalal[0000−0003−0720−7306], Yinglun Zhang[0009−0000−4067−506X], Duru Doğan[0009−0000−4910−1046], Atalay Mert İleri[0009−0002−3610−4963], and Hande Küçük McGinty[0000−0002−9025−5538]

Kansas State University, Manhattan, KS 66502, USA

**Abstract.** The focus on 'food as medicine' is gaining traction in the field of health and several studies conducted in the past few years discussed this aspect of food in the literature. However, very little research has been done on representing the relationship between food and health in a standardized, machine-readable format using a semantic web that can help us leverage this knowledge effectively. To address this gap, this study aims to create a knowledge graph to link food and health through the knowledge graphs' ability to combine information from various platforms focusing on flavonoid contents of food found in the USDA's databases and cancer connections found in the literature. We looked closely at these relationships using KNARM methodology and represented them in machine-operable format. The proposed knowledge graph serves as an example for researchers, enabling them to explore the complex interplay between dietary choices and disease management. Future work for this study involves expanding the scope of the knowledge graph by capturing nuances, adding more related data, and performing inferences on the acquired knowledge to uncover hidden relationships.

## 1 Introduction

Food is important to all living beings as it is essential for growth and energy to perform day-to-day tasks. Food derived from plants are generally rich in bioactive nutrients. These nutrients are beneficial for numerous health benefits, among such nutrients are flavonoids, which are a structurally diverse subclass of polyphenolic compounds with a unit of 2 phenyl chromone produced by plant metabolism. The importance of flavonoids is undervalued in our day-to-day lives. It has a wide variety of advantages, such as several flavonoids help reduce oxidative stress, and they are also proven to reduce inflammation in the human body, which plays an important role in illnesses like cardiovascular diseases. For example, Anthocyanins, flavones, flavonols, and isoflavones are a few flavonoids that have been found to have beneficial effects on asthma. Flavonoids have also been found helpful in cancer; for example, Flavonoids like Luteolin have been found to have anti-cancer abilities [19] [6]. Similarly, the flavonoid quercetin, which is present in apples, onions, and broccoli, exhibited anti-inflammatory activities in a mouse model of allergic airway inflammation[22]. Such properties of Flavonoids have lately sparked the interest of researchers in this field. As, with these applications, it has become essential to know the link of flavonoids and their food sources. However, understanding the intricate relationships between flavonoids and food data can be quite challenging, given the complexity and diversity of these fields. Flavonoids themselves are a group of compounds with various subclasses and structures, making their analysis a complex task. However, knowledge graphs provide a valuable solution by allowing the organization and analysis of complex relationships between different entities and concepts. Constructing a knowledge graph also allows for a comprehensive understanding of the relationships between flavonoids and various aspects of food data by performing inferences on it. The interconnected nature of the knowledge graph enables researchers to identify patterns, dependencies, and correlations that would be difficult to discern from isolated datasets [12]. This includes specific flavonoids, their dietary sources, and potential health effects[12]. Additionally, knowledge graphs are flexible and can be updated and refined as new knowledge and research findings emerge, ensuring that the representation remains up-to-date and accurate.

This paper follows an existing approach of tracing the relationships of food following the FDC Ontology [11] which is related to but different from the FOODON ontology approach [10]. We further generate a knowledge graph (KG) by implementing a standardized language to assist computers in learning the non-linear relationship among both entities. This study is focused to link the different subclasses of the same flavonoids from various food origins to build a machine-operable and AI-ready (Artificial-intelligence-ready) structure for any upcoming downstream analysis and hypothesis generation which further helps in various fields like biomedical research fields.

## 2    Related work

Foods and their flavonoids have been the focus of numerous studies over the years, as they are now recognized as important bioactive chemicals. Flavonoids, which encompass a group of compounds with various subclasses and structures, pose a challenge for analysis due to unstructured data and unorganized way of representing flavonoids. However, the use of knowledge graphs provides a valuable solution in unraveling these complexities.

A study has been conducted which presents the proposal of an ontology to standardize knowledge of fast food and link nutritional data for consumers and experts. The ontology was based on metadata from 21 fast-food establishments, and it discusses the challenges of managing and publishing large amounts of semantically linked fast-food nutritional data and concludes that the ontology can facilitate linked data of nutritional information and provide methods to query the data across heterogeneous source [3]. This research collate and standardize the unstructured nutritional information from different fast food chain and does not relate any data other than limited number of classes in fast food. The main aim of our study is to use the findings of this study to better understand the relationship among flavonoids and health factors to better facilitate the use of this knowledge. Another study conducted by Foodon [10] where it created a well-organized system of food terms and definitions that are all logically connected. The goal is to make it easier to trace where our food comes from and how it moves through the supply chain. The aforementioned comprises a range of food sources derived from both animal and plant origins, as well as distinct categorizations and products. Additionally, it includes a variety of auxiliary attributes, such as preservation methodologies, interaction surfaces, and packaging materials. Although, the study does use food data and its meta data to trace back the production of food by using ontology. However, this study does not expand the course to also relate the nutritional chemicals with the food and where it was produced as it was found in our conducted study. Additionally [6], another study was conducted, which focused on examining the role of flavonoids such as quercetin which has anti-inflammatory properties and is used for a remedy for asthma. However, this study focussed on very limited flavonoids and their food derivation. There are certain similar studies [5][20] found out that flavonoids such as Quercetin and Apigenin may be helpful in reducing the chances of having colon cancer and rectal cancer respectively. Another study also suggested that Luteolin has anti-cancer and anti-tumor abilities and can be helpful in cases of tumor-causing problems [19]. Other study suggested that in human breast cancer cells treated with progestin, apigenin prevents the production of vascular endothelial growth factor mRNA and protein [7]. Most of the studies, as discussed above, in this field has expressed their research interest in the effects of flavonoids and food nutrients on health, leaving a huge gap to convert this data into machine-readable format and uncover hidden and complex relationships using ontologies. Current literature cites the impact of flavonoid-rich diets in preventing some types of cancer[4] [17] [13]. For instance, previous research has revealed intriguing patterns in breast cancer incidence across different regions. For example, Asian countries typically report lower breast cancer rates compared to Western nations. an intriguing trend emerges among Asian women who adhere to a Western dietary pattern. In these cases, the incidence of breast cancer aligns more closely with that of Western women, suggesting a potential link between diet and disease risk. [17] While research in this area remains ongoing, preliminary findings suggest that incorporating flavonoids into conventional cancer treatment regimens may yield positive outcomes. Although more studies are needed to fully understand the extent of their effectiveness, current data points towards the potential therapeutic benefits of flavonoids when used alongside standard anticancer treatments. [13] [4]. Our research focuses on different subclasses and subtypes of flavonoids and their food sources, such as Anthocyanidins, Flavan-3-ols, Flavnaones, Flavones, and Flavonols. They all share a general chemical structure but have different capabilities and different food pro-duced, which is then cleaned and filtered using data processing steps mentioned in the methodology, ultimately leading us to develop ontology[13] .

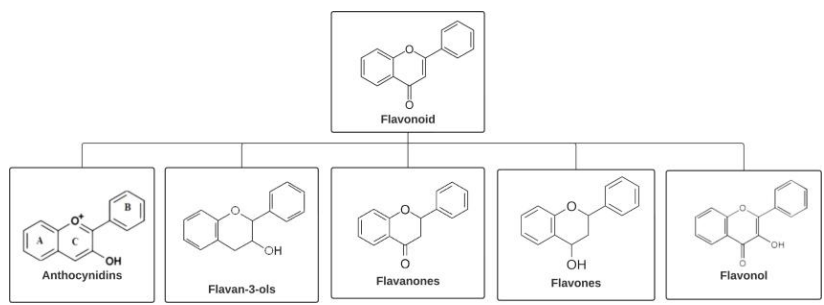

**Fig. 1.** Chemical Structure of Sub classes

## 3    Methodology

### 3.1    Dataset

In this study, the flavonoid datasets provided by the United States Department of Agriculture (USDA) [15] were taken into consideration. The provided datasets contained extensive information regarding diverse flavonoid molecules and their corresponding food sources. This dataset included various information about the flavonoids, such as their origin, states like frozen, seedless, or puree, average content, mean values, and evaluation method. During the course of the research, a significant challenge was encountered in managing the intricate web of unstructured data pertaining to both flavonoids and food products. Some of the main problems were incorrect spelling or different names referencing the same element. In addition to the data from the USDA's database, we used literature that explored relationships among cancers and flavonoid components of food.

To approach this in an agile fashion, we employed the KNARM methodology [18] (see Fig 3), which integrates human expertise and advanced computing capabilities to facilitate the acquisition of knowledge. The process of knowledge representation adopts axioms within a Systematically Deepening Modeling (SDM) framework to establish formal logic-based definitions of concepts [16].

### 3.2    Sub-Language Analysis

The first step of working with the database included cleaning the data and shaping it in a study-usable format. This involved converting datasets from Microsoft Access Database to Microsoft Excel. It was done to achieve access to modern tools and updated support provided by Microsoft to Excel as well as support from tools to develop an ontology, such as Robot [8]. After performing data formatting and structuring, We initiated Sub-language evaluation, which is the primary step within the progress of an ontology that includes recognizing and characterizing the units of data, such as concepts and relationships, that will be used to represent information within the ontology. It is a process of analyzing the language utilized in a specific domain to recognize the key concepts and relationships that are significant to that space. In addition to identifying axioms and entities, we extracted literature data regarding diseases such as cancer and connected it with flavonoids from semantic scholar. In this way, we had a limited sample of studies that connected various food items positive effects on cancer.

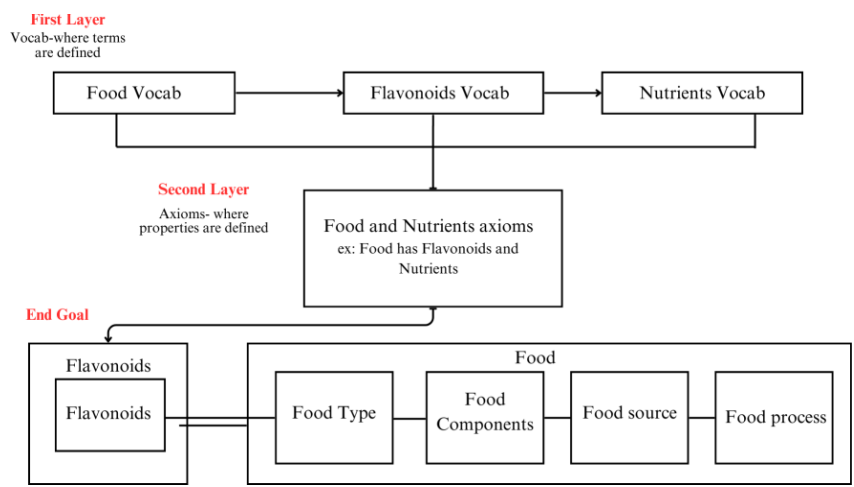

**Fig. 2.** Project Outline

### 3.3   Unstructured Interview

At this step, we start with taking Unstructured Interview, which is conducted to acquire insights from domain experts regarding the data and its functions for augmenting comprehension of the data and its purposes, and informational components used in ontology development. In this study, as we had a good understanding of the data and concepts, hence we acted as domain experts and thereby reduced the pressing demand of bringing outside domain experts, and performed all critical undertakings that needed attention. For instance, manually filtering all the relevant data from the different files of the dataset are required to conduct further studies. We also extracted data from the literature we pulled from the web and integrated the relationships among cancers and food data to our dataset during this step.

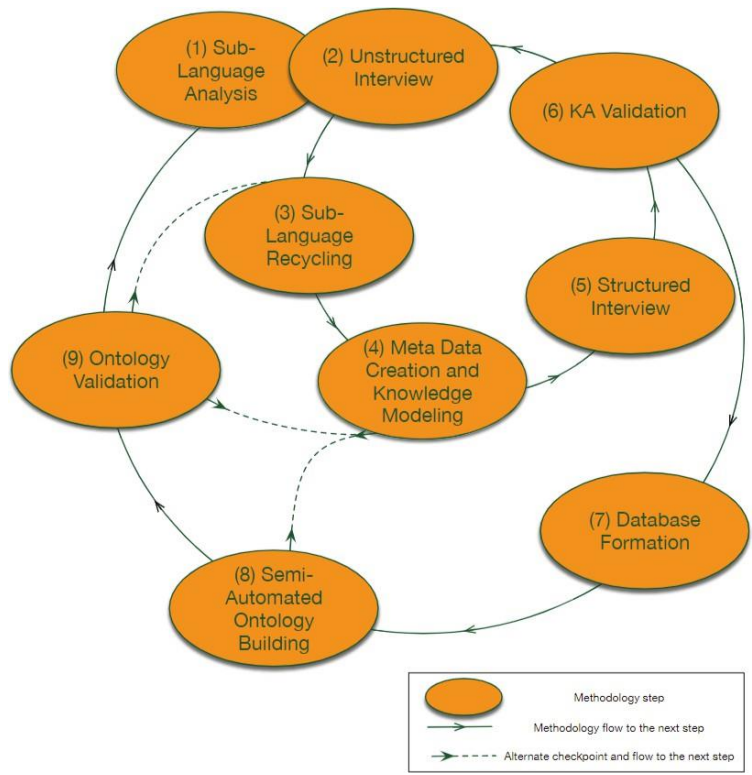

**Fig. 3.** Steps of KNARM Methodology

### 3.4    Sub-Language Recycling (SLR)

After the unstructured interview, we moved to sub-language recycling (SLR) for ontology development. This involves reusing existing terms and definitions from relevant ontologies or vocabularies in the domain. SLR helps avoid duplication, inconsistency, and ambiguity in ontology development. We reviewed relevant ontologies and vocabularies to accomplish SLR. We mapped similar terms to existing ontologies such as Chemical Entities of Biological Interest (ChEBI) [9] and the Compositional Dietary Nutrition Ontology (CDNO) [1] and DOID [21] for disease names for consistency and less redundancy as well as reusing work that is already established and widely used by the community. Our careful term mapping helped avoid duplication; for instance, 'apple' and 'apples' point to the same set of nutrients. However, are termed as different entities in the datasets and recycled vocabularies. Conducting a mapping ensures consistency in ontology development. This saved time and effort and ensured interoperability with other ontologies for knowledge sharing. The recycling process was effective for ontology development. The next step in Ontology development is Metadata Creation and Knowledge Modelling which is the process of creating a representation of knowledge in a specific domain.

### 3.5    Meta Data Creation and Knowledge Modelling

For Metadata Creation and Knowledge Modelling, we integrated cross-referencing mechanisms with Chemical Entities of Biological Interest (ChEBI) and the National Center for Biotechnology Information taxonomies and associated modeling frameworks. By establishing a connection between our ontology terminologies and extant ChEBI entities, various advantages were realized. These benefits include mitigating redundancy, rendering an all-encompassing comprehension of the domain, and enhancing the precision of our ontology. Through the process of mapping to ChEBI and NCBI (National Center for Biotechnology Information) taxonomy, it was established that flavonoids belong to the classification of secondary metabolites of nutrients, which fall within the subclass components present in food products. We also extended the connection of the flavonoids by adding "has associated disease" linkage to flavonoids in conjunction with their respective cancer subtypes. This addition provides insight into the potential effects of flavonoids on the development or management of these particular cancer types.

### 3.6    Structured Interview

As we complete the step of adding metadata and knowledge modeling, we move forward to the Structured Interview, this step included involving a method used to identify important points that may have been missed by knowledge engineers and domain experts during the knowledge acquisition process. Our past experiences with the data allowed us to play the role of domain experts for this work. It eliminated the need to go outside for domain expert's knowledge. This concluded our current step.

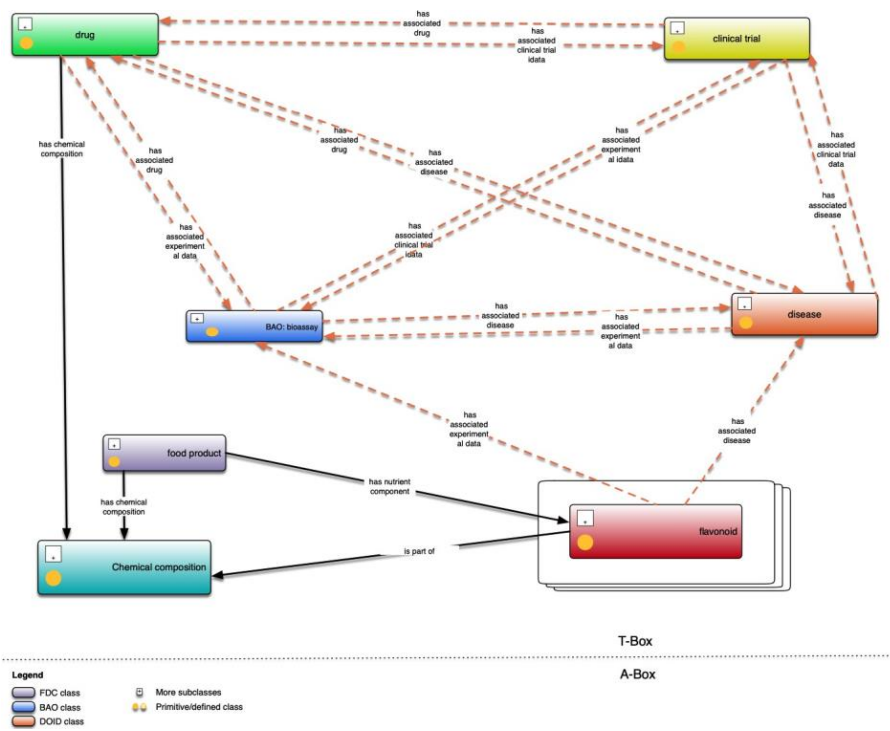

**Fig. 4.** Ontology Structure

## 3.7   Database Formation

During this study, an initial step was the establishment of a database accompanied by the implementation of Knowledge Acquisition (KA) validation processes. For the creation of the database, the Neo4J application was chosen due to its proficient read-and-write capabilities on a graph database. To establish the database, the extant data was extracted and partitioned into distinct CSV sheets. Subsequently, relationships between the various entities were established through the employment of lexical units such as "Parent of" and "Has An ID". We also added the "has associated disease" relationship to flavonoids with their corresponding cancer type, indicating the effect of flavonoids on aiding the given cancer type. As mentioned above, the flavanoid and food data were extracted from the USDA's flavanoid database, while the disease information was generated using the literature that was extracted using semantic scholar [14]. Following a similar perspective as illustrated in Fig. 4 portraying food as comprising a chemical composition that encompasses flavonoids, which are linked to specific diseases. Moreover, this chemical composition contributes to drug formulation, which undergoes clinical trials aimed at treating diseases, thereby supporting our foundational belief that food possesses medicinal properties. This permitted us the opportunity to establish an all-inclusive database, which comprised of the necessary features for enhanced data readability through an interface more conducive to user experience. The utilization of the database facilitated our ability to identify the distinct connections between the entities, which played a vital role in our analysis. It also enables us to add more data to the dataset as needed to continue this study and enhance our knowledge graph, which will be discussed further in the "Future Work" section.

### 3.8   KA Validation

The next step entailing is KA Validation which is a validation of knowledge acquired and accurate representation of the data in the Knowledge model. It involves the assessment of the ontology's quality and consistency through the utilization of both automated and manual approaches. Automated approaches encompass the utilization of reasoners, validators, and metrics tools to identify inaccuracies, inconsistencies, and redundancies within the ontological structure. The manual methods employed in ontology evaluation involve the utilization of domain experts, peer reviews, and user feedback to assess the ontological elements' accuracy, completeness, and usability. For validation purposes, we used the already set up database on Neo4J and formed queries to cross-check our methodologies. For instance, when queried about all the foods in Dairy and Egg Products, to which the response was "Milk, chocolate, fluid, commercial, reduced fat, with added vitamin A and vitamin D" which is what our database reflects, further if we go ahead and query about all the flavonoids Milk contains to which it shows the output of all the flavonoids like "(+)-Catechin" and "(+)-Gallocatechin" and the entire list which is corresponding to our dataset. This query-building system and its output successfully validated our acquired knowledge.

### 3.9   Semi-Automated Ontology Building

During this phase, we expanded our dataset with information concerning the association between flavonoids and diseases such as breast cancer and colon cancer. Furthermore, we integrated literature data obtained through Semantic Scholar, establishing connections between flavonoids and cancer. This approach enabled more nuanced mapping of these entities to existing ones, generating a new systematic approach for connecting food to disease and drug to disease using "chemical composition components" as their common association point which may allow more in-depth analyses of flavonoid data and hypotheses generation to evaluate their impact on various health metrics.

After the completion of organized information building and verification, another step included the creation of ontologies through semi-automated tools. In this respect, we selected for [8] ROBOT, a command line tool that encouraged the ontology-building process. Robot required an organized CSV file as an input layout together with prefixes, IRI's, and output location. By utilizing ROBOT, we were able to rapidly produce the ontology, avoiding any potential errors which will emerge due to manual input. This semi-automated approach too permitted us to extend the effectiveness of the ontology-building process and diminished the probability of irregularity or redundancy.

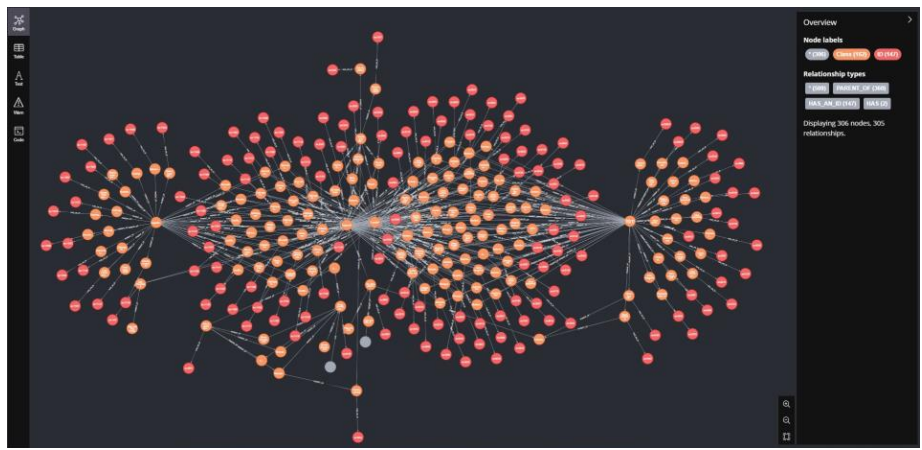

**Fig. 5.** Database representation in graph format

By analyzing data in the CSV file format, we were able to gain a thorough understanding of the robot's behavior, as well as the underlying reasons for its operation. For instance, It is important to note that when programming in Robot, attention must be given to case sensitivity and accurate spelling. Once the structure of the CSV file aligns with the Robot template file, we followed a 3 layer structure of the model (see Fig. 2) where the first layer consisted of vocabulary, the second layer consisted of merging the files to make axioms and lastly in the third level we merge the flavonoids files into the foods files.

### 3.10   Ontology Validation

As for Ontology Validation, It lies outside the scope of this paper as we are currently and continuously working on validating our ontology.

## 4   Conclusion

The objective of this research was to uncover existing correlations and associations between flavonoids and various consumable items derived from diverse origins and sources and their connection to aiding different types of cancers.

Depicting these associations in ontology framework by utilizing the dataset maintained by the United States Department of Agriculture (USDA) and adding data for disease connections we generated a new modelling paradigm that allowed us to connect drugs-to-food-to-disease using 'chemical composition component' as their connection point as seen is Figure 4. This connection clearly creates a new path for connecting drugs to food for through their disease connection too which may allow further hypotheses generation for "food as medicine" approach. The research was motivated by the need to develop a more profound comprehension of the complex interrelations between flavonoids, food items and their relation with various types of cancers, ultimately facilitating more efficacious evaluation and decision-making practices within the realms of nutrition, health, and food science. The present study followed steps of data collection and structuring, ontology development, and database creation and validation. The outcomes of this study may have consequential implications for the formulation of novel dietary directives and the recognition of fresh therapeutic objectives in the domain of nutraceutical and medical inquiry.

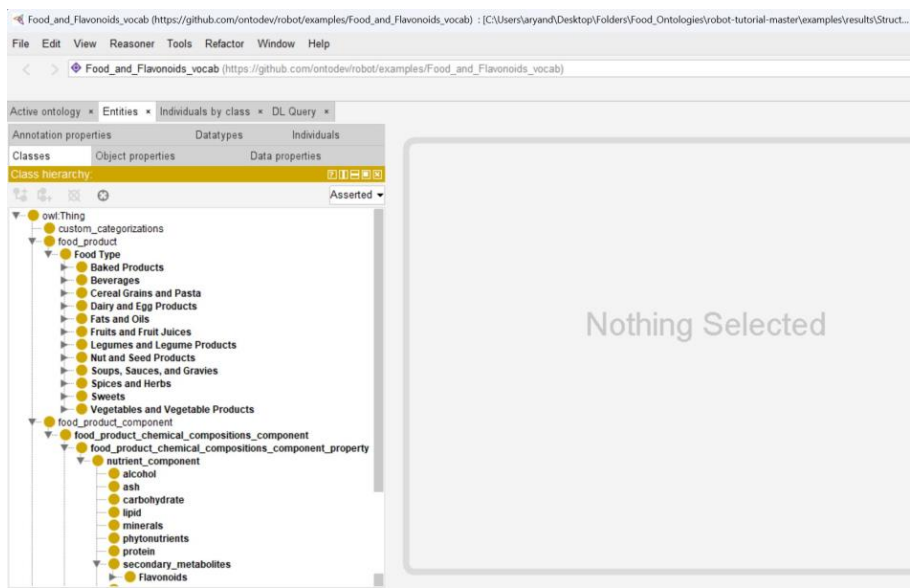

**Fig. 6.** Owl file made using Robot and Protégé

## 5 Future Work

As we look to progress in un-nesting complex relationships in the data through ontology development and converting it into machine-readable format. It can not be denied that ontology development is tedious and human-intensive work. To reduce the dependency on domain experts to develop ontology from data of various fields, involving Large Language Models (LLM) for the development process by using OLIVE (Ontology Learning with Integrated Vector Embeddings)[2] becomes important. We are actively working to finalize our OLIVE workflow, which will enable users to generate knowledge knowledge graphs. This would help in reducing human involvement and expediting the ontology development timeline.

# 6    Acknowledgements

Parts of this research was conducted via support from NIH grant K-INBRE P20 GM103418.

## 6.1    Supplemental Material

Further documentation and files can be found on the page:
https://github.com/koncordantlab/Flavonoid-Fusion.git